\documentclass[runningheads]{llncs}
\usepackage{booktabs}
\usepackage{graphicx}
\usepackage{multirow}
\usepackage{amssymb}
\usepackage{pythonhighlight}
\usepackage{amsmath}
\usepackage{wrapfig}
\usepackage{booktabs}
\usepackage{diagbox}
\usepackage{multirow}
\usepackage{gensymb}
\usepackage{ulem}
\usepackage{bbding}
\usepackage{float}
\usepackage{caption}

\definecolor{ao}{rgb}{0.01, 0.75, 0.24}
\usepackage[colorlinks,linkcolor=red, anchorcolor=blue, citecolor=blue, urlcolor=magenta]{hyperref}

\begin{document}
%
\title{Universal Semi-Supervised Learning for Medical Image Classification}

\titlerunning{Universal Semi-Supervised Classification}
\author{Lie Ju\inst{1, 2, 3},
Yicheng Wu\inst{3, 4},
Wei Feng\inst{1,2,3},
Zhen Yu\inst{1,2,3}, \\
Lin Wang\inst{1, 3, 5},  
Zhuoting Zhu\inst{6}, \and 
Zongyuan Ge\inst{1,2,3(}\Envelope\inst{)}}

\authorrunning{L. Ju et al.}
%
\institute{Monash-Airdoc Research, Monash University, Melbourne, Australia \and
eResearch Centre, Monash University, Melbourne, Australia \and
Monash Medical AI Group, Monash University, Melbourne, Australia \and
Faculty of Information Technology, Monash University, Melbourne, Australia \and
Harbin Engineering University, Harbin, China \and
Centre for Eye Research Australia, Melbourne University, Melbourne, Australia
\\ \url{https://www.monash.edu/mmai-group}
\\
\email{Lie.Ju1@monash.edu}, \email{zongyuan.ge@monash.edu}}

\maketitle              
\begin{abstract}
Semi-supervised learning (SSL) has attracted much attention since it reduces the expensive costs of collecting adequate well-labeled training data, especially for deep learning methods. However, traditional SSL is built upon an assumption that labeled and unlabeled data should be from the same distribution \textit{e.g.,} classes and domains. However, in practical scenarios, unlabeled data would be from unseen classes or unseen domains, and it is still challenging to exploit them by existing SSL methods. Therefore, in this paper, we proposed a unified framework to leverage these unseen unlabeled data for open-scenario semi-supervised medical image classification. We first design a novel scoring mechanism, called dual-path outliers estimation, to identify samples from unseen classes. Meanwhile, to extract unseen-domain samples, we then apply an effective variational autoencoder (VAE) pre-training. 
After that, we conduct domain adaptation to fully exploit the value of the detected unseen-domain samples to boost semi-supervised training. We evaluated our proposed framework on dermatology and ophthalmology tasks. Extensive experiments demonstrate our model can achieve superior classification performance in various medical SSL scenarios. The code implementations
are accessible at:  \url{https://github.com/PyJulie/USSL4MIC}.


\end{abstract}

\section{Introduction}

Training a satisfied deep model for medical classification tasks remains highly challenging due to the expensive costs of collecting adequate high-quality annotated data. Hence, semi-supervised learning (SSL)~\cite{laine2016temporal,berthelot2019mixmatch,sohn2020fixmatch,lee2013pseudo,liu2020semi} becomes a popular technique to exploit unlabeled data with only limited annotated data. Essentially, most existing SSL methods are based on an assumption that labeled and unlabeled data should be from the same close-set distribution and neglect the realistic scenarios. However, in practical clinical tasks (\textit{e.g.,} skin lesion classification), unlabeled data may contain samples from unknown/open-set classes which do not present in the training set, leading to sub-optimal performance. 

\begin{figure}[t]
    \centering
    \includegraphics[width=12cm]{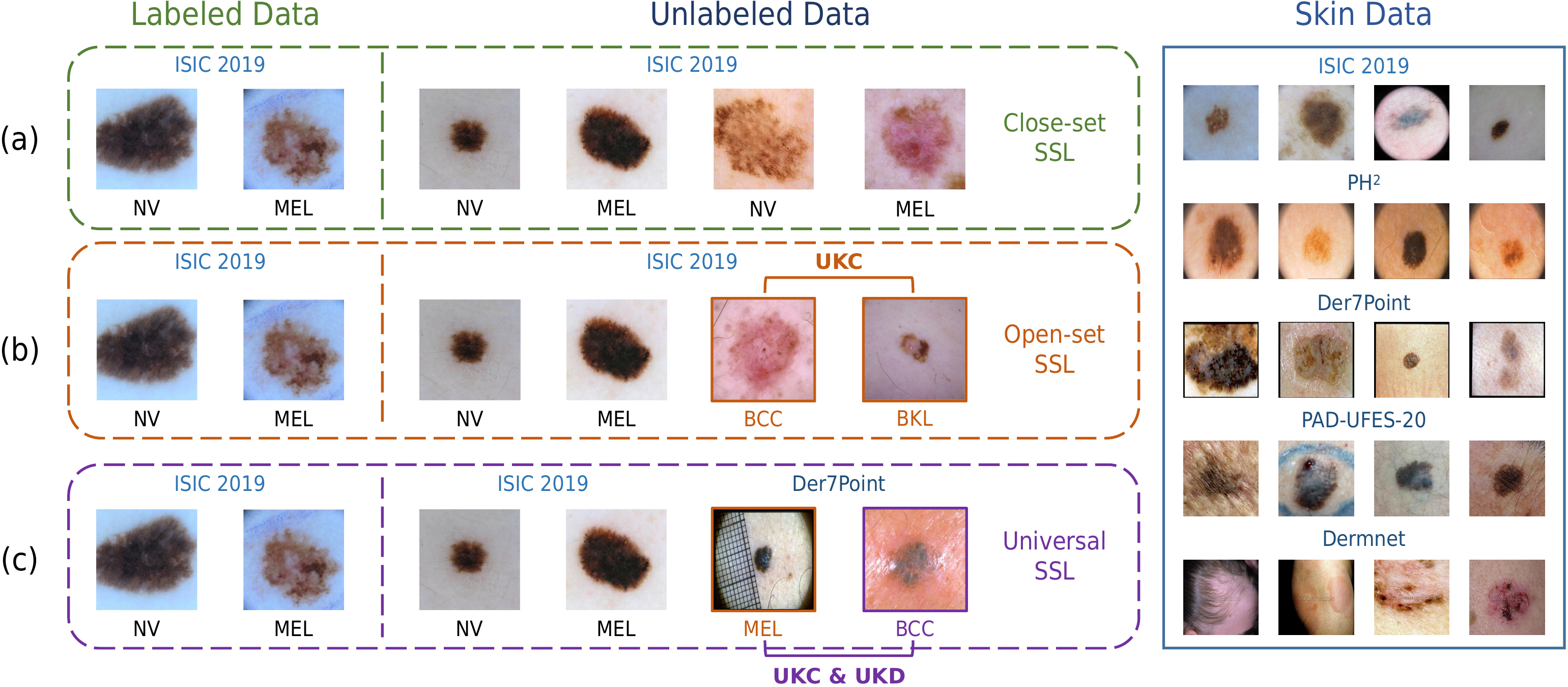}
    \caption{Problem illustration. (a) Close-set SSL. The samples in the labeled and unlabeled data share the same classes and are collected under the same environment, \textit{i.e.,} dermatoscopes. (b) Open-set SSL. There are unknown classes (UKC) in the unlabeled data, \textit{e.g.,} BCC and BKL. (c) Universal SSL. In addition to the unknown classes, the samples in the unlabeled data may come from other unknown domains (UKD), \textit{e.g.,} samples from other datasets with different imaging and condition settings.}
    \label{fig:problem definition}
\end{figure}

We further illustrate this problem in Fig.~\ref{fig:problem definition}. Specifically, Fig.~\ref{fig:problem definition}-(a) shows a classic close-set SSL setting: the labeled and unlabeled data from ISIC 2019 dataset~\cite{isic} share the same classes, \textit{i.e.,} melanocytic nevus (NV) and melanoma (MEL). Fig.~\ref{fig:problem definition}-(b) shows a condition of \textbf{Open-set SSL}, where novel classes of basal cell carcinoma (BCC) and benign keratosis (BKL) are introduced. Recent works for Open-set SSL~\cite{saito2021openmatch,lee2022contrastive} mainly focused on identifying those outliers during the model training. Unlike them, here, we further consider a more realistic scenario that also greatly violates the close-set assumption posted above, as shown in Fig.~\ref{fig:problem definition}-(c). The unlabeled data may share the same classes but come from quite different domains, \textit{e.g.,} MEL from Der7point dataset~\cite{esteva2017dermatologist}, which contains clinical images in addition to dermoscopic images. Meanwhile, there are some unknown novel classes, \textit{e.g.,} BCC from Der7point dataset, leading to both \textbf{seen/unseen class} and \textbf{domain} mismatch.

To handle this mismatch issue, \textit{Huang} et al.~\cite{huang2021universal} proposed CAFA to combine the open-set recognition (OSR) and domain adaptation (DA), namely \textbf{Universal SSL}. Specifically, they proposed to measure the possibility of a sample to be unknown classes (UKC) or unknown domains (UKD), which are leveraged to re-weight the unlabeled samples. 
The domain adaptation term can adapt features from unknown domains into the known domain, ensuring that the model can fully exploit the value of UKD samples. The effectiveness of CAFA relies heavily on the detection of open-set samples where the proposed techniques always fail to generalize on medical datasets. 
For medical images, such as skin data, UKC and UKD samples can be highly inseparable (\textit{e.g.,} MEL in Fig.~\ref{fig:problem definition} - (a) vs. BCC in Fig.~\ref{fig:problem definition} - (b)), particularly when training with limited samples in a semi-supervised setting.

Therefore, in this work, we propose a novel universal semi-supervised framework for medical image classification for both class and domain mismatch. Specifically, to measure the possibility of an unlabeled sample being UKC, we propose a dual-path outlier estimation technique, to measure the possibility of an unlabeled sample being UKC in both feature and classifier levels using prototypes and prediction confidence. In addition, we first present a scoring mechanism to measure the possibility of an unlabeled sample being UKD by pre-training a Variational AutoEncoder (VAE) model, which is more suitable for medical image domain separation with less labeled samples required. With the detected UKD samples, we applied domain adaptation methods for feature matching for different domains. After that, the labeled and unlabeled samples (including feature-adapted UKD samples) could be optimized using traditional SSL techniques. 

Our contributions can be summarized as: (1) We present a novel framework for universal semi-supervised medical image classification, which enables the model to learn from unknown classes/domains using open-set recognition and domain adaptation techniques. (2) We propose a novel scoring mechanism to improve the reliability of the detection of outliers from UKC/UKD for further unified training. (3) Experiments on datasets with various modalities demonstrate our proposed method can perform well in different open-set SSL scenarios.

\section{Methodology}
\subsection{Overview}
\begin{figure}[t]
    \centering
    \includegraphics[width=12cm]{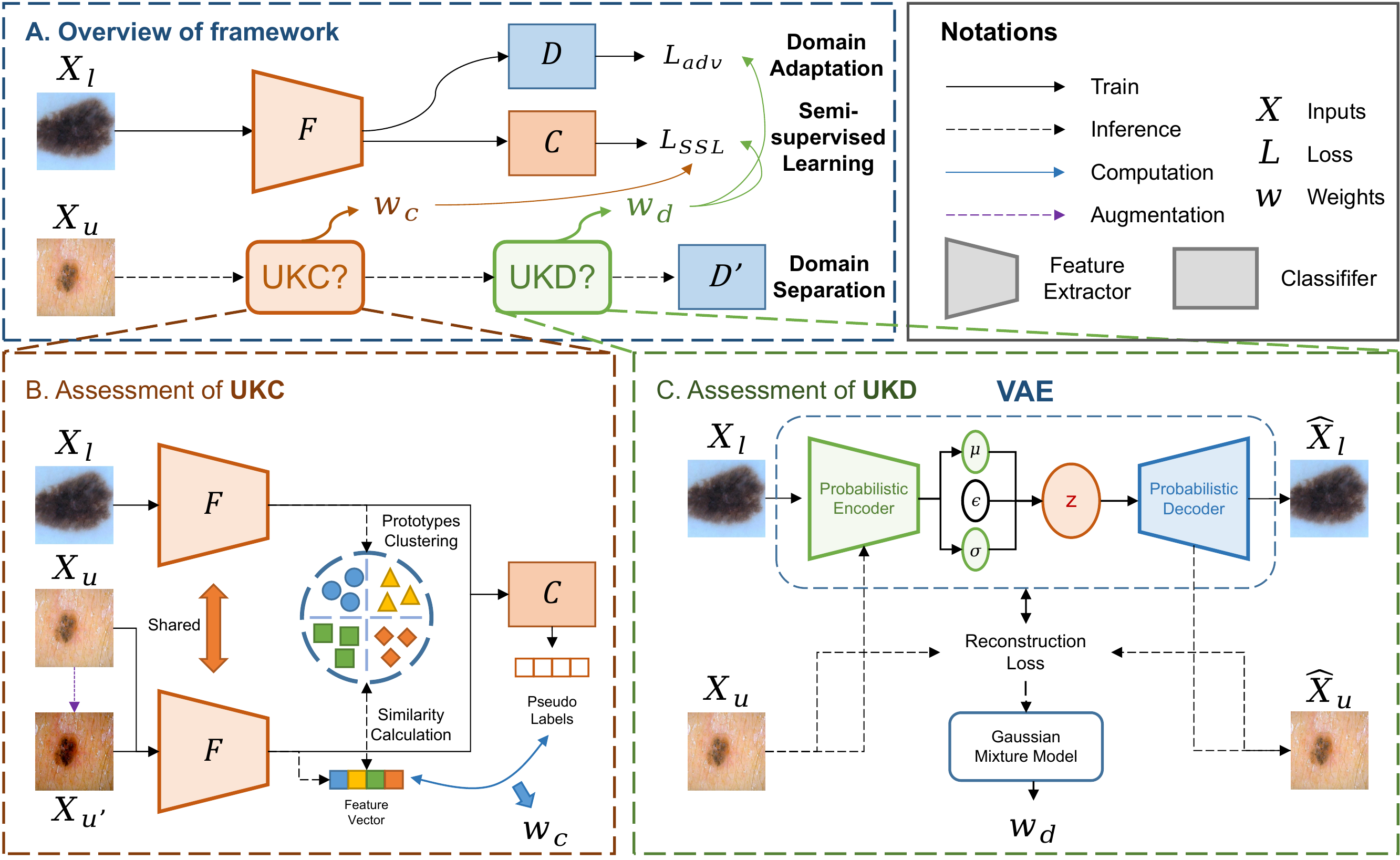}
    \caption{The overview of our proposed framework.}
    \label{fig:framework}
\end{figure}

The overview of our proposed framework is shown in Fig.~\ref{fig:framework}. The framework mainly contains a feature extractor $\mathcal{F}$, an adversarial discriminator $\mathcal{D}$, a multi-class classifier $\mathcal{C}$, and a non-adversarial discriminator $\mathcal{D'}$. The feature extractor encodes the inputs $\mathcal{X}$ into features $\mathcal{V}$. The multi-class classifier $\mathcal{C}$ outputs the predictions of exact diseases. The non-adversarial discriminator predicts the possibility of an instance from unlabeled data to be UKD. The adversarial discriminator conducts feature adaptation on the samples from known and detected unknown domains. To summarize, our target is to score unlabeled samples from unseen classes/domains for further training using SSL and domain adaptation.

\subsection{Dual-path Outlier Estimation}
Recent open-set SSL methods~\cite{saito2021openmatch,huang2021trash} mainly focus on the detection of UKC samples, which is known as the OSR task. Those outliers will be removed during the training phase. In this section, we propose a novel OSR technique namely Dual-path Outlier Estimation (DOE) for the assessment of UKC based on both feature similarities and confidence of classifier predictions. Formally, given labeled samples $\mathcal{X}_{l}$, we first warm-up the model with standard cross-entropy loss.
Unlike CAFA~\cite{huang2021universal}, which computes instance-wise feature similarity, we argue that samples from known classes should have closer distances to the centric representations, e.g, prototypes, than outliers. The prototypes of a class can be computed as average outputs of its corresponding samples $x_{l,i} \in \mathcal{X}_{l}$:
\begin{equation}
    \textbf{v}_{l, c_{j}} = \frac{\sum_{i=1, x_{l, i}\in \mathcal{X}_{l, c_{j}}}^{N_{c_{j}}} \mathcal{F}(x_{l, i})}{N_{c_{j}}},
\end{equation}
where $N_{c_{j}}$ denotes the number of instances of class $j$ and $\textbf{v}_{c_{j}}$ is a vector with the shape of 1$\times$D after the average global pooling layer. Then, the feature similarity of an instance $x_{u,i} \in \mathcal{X}_{U}$ to each known class can be calculated as:
\begin{equation}
    \textbf{d} = \{d_{i, x_{i}\in c_{j}}\}_{j=1}^{N_{c_{j}}} = \{\left\|\mathcal{F}(x_{u,i|c_{j}}) - \textbf{v}_{l, c_{j}}\right\|_{2}\}_{j=1}^{N_{c_{j}}}.
\end{equation}
We can assume that if a sample is relatively far from all class-specific prototypes, it should have a larger average value $d_{avg}$ of distance \textbf{d} and can be considered a potential outlier~\cite{ju2022flexible,ming2022exploit,ye2021towards}. Then, we perform strong augmentations on unlabeled inputs and generate two views $x_{u'_{i,1}}$ and $x_{u'_{i,2}}$, which are also subsequently fed into the pre-trained networks and obtain the predictions $\textbf{p}_{u_{i,1}}$ and $\textbf{p}_{u_{i,2}}$. Inspired by agreement maximization principle~\cite{sindhwani2005co}, a sample to be outliers or not can be determined by the consistency of these two predictions:
\begin{equation}
    p_{ood}(x_{i}) = |\textrm{max}\ (\textbf{p}_{u_{i,1}}) - \textrm{max}\ (\textbf{p}_{u_{i,2}})|.
\end{equation}
Finally, we combine the prototype-based and prediction-based scores:
\begin{equation}
    w_{u,c} = 1 - \sigma(d_{avg}\cdot p_{ood}),
\end{equation}
where $\sigma$ is a normalization function that maps the original distribution into (0,1].

\subsection{Class-agnostic Domain Separation}
Although our proposed DOE can help detect potential UKC samples, an obvious issue is that the domain difference can easily disturb the estimation of UKC, \textit{e.g.,} UKD samples have larger distances to the prototypes of known domains. Different from the detection of UKC, distinguishing UKD samples in the unlabeled data is less difficult since there is more environmental gap among different domains such as imaging devices, modalities, or other artifacts. To this end, we adopt the VAE which is agnostic to the supervised signal and can pay more attention to the global style of the domain~\cite{sun2020conditional}. Formally, VAE consists of an encoder $g(\cdot)$ and a decoder $f(\cdot)$, where the encoder compresses high-dimensional input features $x_{i}$ to a low-dimension embedding space and the decoder aims to reconstruct from that by minimizing the errors:
\begin{equation}
    \mathcal{L}_{re} = \left\|x_{i} - f(g(x_{i}))\right\|_{2}^{2}.
\end{equation}
In our scenario, we pre-train a VAE model using labeled data and evaluated it on the unlabeled data to obtain the reconstruction errors. Then, we fit a two-component Gaussian Mixture Model (GMM) using the Expectation-Maximization algorithm, which has flexibility in the sharpness of distribution and is more sensitive to low-dimension distribution~\cite{lidividemix}, \textit{i.e.,} the reconstruction losses $\mathcal{L}_{re}$. For each sample $x_{i}$, we have its posterior probability as $w_{d,i}$ for domain separation. With known domain samples from labeled data (denoted as $y_{l,d,i} = 0)$ and the possibility of UKC samples from unlabeled data (denoted as $y_{u,d,j} = w_{d,j}$), we optimize a binary cross-entropy loss for non-adversarial discriminator $\mathcal{D}'$:

\begin{equation}
    \mathcal{L}_{dom} = -\frac{1}{N_{l}}\sum_{i=1}^{N_{l}}{1-\log(\hat{y}_{l,d,i})}-\frac{1}{N_{u}}\sum_{j=1}^{N_{u}}w_{d,j}\cdot\log(\hat{y}_{u,d,j})-(1-w_{d,j})\cdot(\log(1-\hat{y}_{u,d,j})).
\end{equation}

\subsection{Optimization with Adversarial Training \& SSL}
To make a domain adaptation for distinguished unknown domains, we adopt a regular adversarial training manner~\cite{cao2018partial}, with the labeled data treated as the target domain and the unlabeled data as the source domain. Note that we use two weights $w'_{d,u}$ from the non-adversarial discriminator and $w_{c,u}$ from DOE to determine which samples to adapt. The adversarial loss can be formulated as:
\begin{equation}
    \mathop{\max} \limits_{\theta_{F}} \mathop{\min} \limits_{\theta_{D}} \mathcal{L}_{adv} = -(1-y_{t})\cdot\log(1-D(F(\mathcal{X}_{l}))) - y_{s}\cdot w'_{u,d}\cdot w_{u,c}\cdot\log D(F(\mathcal{X}_{u})),
\end{equation}
where $\theta$ denotes the parameters of the specific module and $y_{s}$ = 1 and $y_{t}$ = 0 are the initial domain labels for the source domain and target domain.
Then, we can perform unified training from the labeled data and selectively feature-adapted unlabeled data under weights controlled. The overall loss can be formulated as:
\begin{equation}
    \mathcal{L}_{overall} = \mathcal{L}_{CE}(\mathcal{X}_{l}) - \alpha \cdot \mathcal{L}_{adv}(\mathcal{X}_{u}|w'_{u,d},w_{u,c}) + \beta \cdot \mathcal{L}_{SSL}(\mathcal{X}_{u}|w_{u,c}),
\end{equation}
where $\alpha$ and $\beta$ are coefficients. For the semi-supervised term $\mathcal{L}_{SSL}$, we adopt $\Pi$-model~\cite{laine2016temporal} here. Thus, we can perform a global optimization to better utilize the unlabeled data with the class/domain mismatch.

\section{Experiments}
\subsection{Datasets \& Implementation Details}

\textbf{Dermatology} For skin lesion recognition, we use four datasets to evaluate our methods: ISIC 2019~\cite{isic}, PAD-UFES-20~\cite{pacheco2020pad}, Derm7pt~\cite{kawahara2018seven} and Dermnet~\cite{Dermnet}. The statistics of four datasets can be found in our supplementary documents. The images in ISIC2019 are captured from dermatoscopes. The images in PAD-UFES-20 and Dermnet datasets are captured from a clinical scenario, where Derm7pt dataset contains both. 
Firstly, we divide ISIC 2019 dataset into 4 (NV, MEL, BCC, BKL) + 4 (AK, SCC, VASC, DF) classes as known classes and unknown classes respectively. We sample 500 instances per class from known classes to construct the labeled datasets. Then we sample 250 / 250 instances per class from known classes to construct validation datasets and test datasets, We sample 30\% close-set samples and all open-set samples from the left 17,331 instances to form the unlabeled dataset. For the other three datasets, we mix each dataset with ISIC 2019 unlabeled dataset, to validate the effectiveness of our proposed methods on training from different unknown domains.

\noindent \textbf{Ophthalmology}
We also evaluate our proposed methods on in-house fundus datasets, which were collected from regular fundus cameras, handheld fundus cameras, and ultra-widefield fundus imaging, covering the field of view of 60\degree, 45\degree, and 200 \degree respectively. We follow~\cite{ju2021synergic,ju2021leveraging} and take the diabetic retinopathy (DR) grading with 5 sub-classes (normal, mild DR, moderate DR, severe DR, and proliferative DR) as known classes. We sample 1000 / 500 / 500 instances per class to construct the training/validation/test dataset. The samples with the presence of age-related macular degeneration (AMD) which have similar features to DR, are introduced as 4 unknown classes (small drusen, big drusen, dry AMD, and wet AMD). Please refer to our supplementary files for more details.

\noindent \textbf{Implementation Details}
All Skin images are resized into 224$\times$224 pixels and all fundus images are resized into 512$\times$512 pixels. We take ResNet-50~\cite{he2016deep} as our backbones for the classification model and VAE training. We use $\Pi$-model~\cite{laine2016temporal} as a SSL regularizer. We warm up the model using exponential rampup~\cite{laine2016temporal} with 80 out of 200 epochs, to adjust the coefficients of adversarial training $\alpha$ and SSL $\beta$. We use SGD optimizer with a learning rate of 3$\times10^{-4}$ and a batch size of 32. Some regular augmentation techniques are applied such as random crop, and flip, with color jitter and gaussian blur as strong augmentations for the assessment of UKC. For a fair comparison study, we kept all basic hyper-parameters such as augmentations, batch size, and learning rate the same on comparison methods.

\subsection{Comparison Study}

\begin{table}[t]
\caption{The comparative results on skin datasets (5-trial average accuracy\%).}
\centering
\begin{tabular}{c|c|cccc}
\hline
\hline
                               & Datasets     & \multicolumn{1}{c|}{ISIC 2019}     & \multicolumn{1}{c|}{Derm7pt} & \multicolumn{1}{c|}{PAD-UFES-20} & Dermnet \\ \hline
                Supervised     & ERM & \multicolumn{4}{c}{69.2 ($\pm$ 0.89)}                                                             \\ \hline
\multirow{6}{*}{Close-set SSL} & PL~\cite{lee2013pseudo}         & \multicolumn{1}{c|}{69.4 ($\pm$ 0.65)}  & \multicolumn{1}{c|}{70.1 ($\pm$ 0.10)}  & \multicolumn{1}{c|}{70.2  ($\pm$ 0.83)}      & 66.9 ($\pm$ 0.34)      \\
                               & PI~\cite{laine2016temporal}     & \multicolumn{1}{c|}{70.2 ($\pm$ 0.33)}  & \multicolumn{1}{c|}{ \underline{70.7} ($\pm$ 0.64)}      & \multicolumn{1}{c|}{ \underline{70.3} ($\pm$ 0.22)}      &  \underline{69.6} ($\pm$ 0.87)      \\
                               & MT~\cite{tarvainen2017mean}     & \multicolumn{1}{c|}{69.6 ($\pm$ 0.45)} & \multicolumn{1}{c|}{70.1 ($\pm$ 0.20)}      & \multicolumn{1}{c|}{68.7 ($\pm$ 0.36)}      & 65.1 ($\pm$ 1.43)      \\
                               & VAT~\cite{miyato2018virtual}    & \multicolumn{1}{c|}{70.3 ($\pm$ 0.29)}  & \multicolumn{1}{c|}{69.9 ($\pm$ 0.23)}      & \multicolumn{1}{c|}{69.2 ($\pm$ 0.48)}      &  \underline{69.6} ($\pm$ 0.59)      \\
                               & MM~\cite{berthelot2019mixmatch} & \multicolumn{1}{c|}{59.4 ($\pm$ 2.23)} & \multicolumn{1}{c|}{60.2 ($\pm$ 0.55)}  & \multicolumn{1}{c|}{60.3 ($\pm$ 1.65)}  & 41.7 ($\pm$ 3.97)      \\
                               & FM~\cite{sohn2020fixmatch}      & \multicolumn{1}{c|}{59.8 ($\pm$ 1.88)} & \multicolumn{1}{c|}{65.1 ($\pm$ 0.63)}  & \multicolumn{1}{c|}{63.0 ($\pm$ 1.01)}  & 53.2 ($\pm$ 1.87)     \\ \hline
\multirow{3}{*}{Open-set SSL}  & UASD~\cite{chen2020semi}        & \multicolumn{1}{c|}{70.0 ($\pm$ 0.47)} & \multicolumn{1}{c|}{67.5 ($\pm$ 1.30)}      & \multicolumn{1}{c|}{68.5 ($\pm$ 1.01)}      & 61.2 ($\pm$ 2.85)      \\
                               & DS3L~\cite{guo2020safe}         & \multicolumn{1}{c|}{69.3 ($\pm$ 0.87)} & \multicolumn{1}{c|}{69.8 ($\pm$ 0.55)}      & \multicolumn{1}{c|}{68.9 ($\pm$ 0.76)}      & 65.3 ($\pm$ 1.14)      \\
                               & MTCF~\cite{yu2020multi}         & \multicolumn{1}{c|}{65.4 ($\pm$ 1.99)} & \multicolumn{1}{c|}{69.2 ($\pm$ 0.89)}      & \multicolumn{1}{c|}{66.3 ($\pm$ 0.76)}      &  66.8 ($\pm$ 1.01)     \\ 
                               & T2T~\cite{huang2021trash}         & \multicolumn{1}{c|}{60.2 ($\pm$ 0.23)} & \multicolumn{1}{c|}{61.7 ($\pm$ 0.15)}      & \multicolumn{1}{c|}{60.3 ($\pm$ 0.21)}      &  60.9 ($\pm$ 0.10)     \\ 
                               & OM~\cite{saito2021openmatch}         & \multicolumn{1}{c|}{70.1 ($\pm$ 0.29)} & \multicolumn{1}{c|}{69.6 ($\pm$ 0.54)}      & \multicolumn{1}{c|}{65.8 ($\pm$ 0.46)}      &  65.4 ($\pm$ 0.54)     \\ \hline
\multirow{2}{*}{Universal SSL} & CAFA~\cite{huang2021universal}  & \multicolumn{1}{c|}{68.3 ($\pm$ 1.08)} & \multicolumn{1}{c|}{63.3 ($\pm$ 1.02)}      & \multicolumn{1}{c|}{65.3 ($\pm$ 1.70)}      &  63.3 ($\pm$ 1.88)     \\
                               & Ours       & \multicolumn{1}{c|}{\textbf{71.1} ($\pm$ 1.31)}      & \multicolumn{1}{c|}{\textbf{70.9} ($\pm$ 1.01)}      & \multicolumn{1}{c|}{\textbf{70.8} ($\pm$ 0.98)}      &  \underline{69.6} ($\pm$ 1.28)     \\ \hline \hline
\end{tabular}
\label{table_skin}
\end{table}

\begin{table}[t]
\centering
\caption{The comparative results on fundus datasets (5-trial average accuracy\%).}
\begin{tabular}{c|c|ccc}
\hline
\hline
                               & Datasets & \multicolumn{1}{c|}{Regular} & \multicolumn{1}{c|}{Handheld} & UWF \\ \hline
Supervised                     & ERM      & \multicolumn{3}{c}{70.96 ($\pm$ 0.98)}                                              \\ \hline
Close-set SSL                  & PI~\cite{laine2016temporal}       & \multicolumn{1}{c|}{\underline{75.01} ($\pm$ 0.86)}        & \multicolumn{1}{c|}{\underline{74.51} ($\pm$ 1.20)}        &  73.85 ($\pm$ 1.65)   \\ \hline
Open-set SSL                   & UASD~\cite{chen2020semi}     & \multicolumn{1}{c|}{73.25 ($\pm$ 2.51)}        & \multicolumn{1}{c|}{73.05 ($\pm$ 1.35)}        &  \underline{73.96} ($\pm$ 1.77)   \\ \hline
\multirow{2}{*}{Universal SSL} & CAFA~\cite{huang2021universal}     & \multicolumn{1}{c|}{72.00 ($\pm$ 2.34)}        & \multicolumn{1}{c|}{73.52 ($\pm$ 1.76)}        & 65.21 ($\pm$ 4.21)    \\
                               & Ours     & \multicolumn{1}{c|}{\textbf{77.55} ($\pm$ 2.85)}        & \multicolumn{1}{c|}{\textbf{78.02} ($\pm$ 2.01)}        & \textbf{74.10} ($\pm$ 2.36)    \\ \hline \hline
\end{tabular}
\label{table_fundus}
\end{table}

\noindent \textbf{Performance on Skin Dataset}
As shown in Table~\ref{table_skin}, we have compared our proposed methods with existing SSL methods, which are grouped with respect to different realistic scenarios. It can be seen that our proposed methods achieve competitive results on all datasets with different domain shifts. An interesting finding is that existing methods in open-set SSL, such as MTCF~\cite{yu2020multi}, do not work well in our settings. This is because these methods rely heavily on the estimation and removal of UKC samples. However, the OSR techniques used, which are designed for classical image classification, are not applicable to medical images.

\noindent \textbf{Performance on Fundus Dataset}
Table~\ref{table_fundus} reports the comparative results on fundus datasets. We select one technique for comparison study that shows the best results in each group for different SSL scenarios. Our proposed methods achieve the best performance over other competitors and significant improvements over baseline models in all settings. Unlike the results on the skin dataset, CAFA achieves satisfactory results except for UWF. UASD improves the performance over the baseline ERM model. This is probably because DR and AMD share similar features or semantic information such as hemorrhage and exudates, which can well enhance the feature learning~\cite{ju2021synergic}.

\noindent \textbf{Novel Domain Detection}
As we claim that our proposed scoring mechanism can well identify UKD samples, we perform experiments on the unknown domain separation using different techniques. 
Our proposed CDS scoring mechanism achieves the best results for unknown domain separation. Moreover, we can find that UKD samples are also sensitive to prototype distances, \textit{e.g.,} with a high AUC of 80.79\% in terms of Dermnet (DN), which confirms the importance and necessity of disentangling the domain information for the detection of UKC.

\begin{table}[t]
  \centering
  \begin{minipage}[c]{0.52\textwidth}
    \captionof{table}{Domain separation AUC\%.\label{table_domain_separation}}
    \centering
    \begin{tabular}{c|cccc}
      \hline  \hline
      Datasets        & \multicolumn{1}{c|}{ISIC}  & \multicolumn{1}{c|}{D7pt} & \multicolumn{1}{c|}{PAD}   & DN \\ \hline
      Confidence        & \multicolumn{1}{c|}{59.16} & \multicolumn{1}{c|}{61.94}  & \multicolumn{1}{c|}{63.01} & 68.41   \\ \hline
      Pertur.~\cite{huang2021universal} & \multicolumn{1}{c|}{61.12} & \multicolumn{1}{c|}{62.55}  & \multicolumn{1}{c|}{61.32} & 60.17   \\ \hline
      VAE               & \multicolumn{1}{c|}{66.02} & \multicolumn{1}{c|}{66.32}  & \multicolumn{1}{c|}{63.28} & 79.09   \\ \hline
      OVA-Net~\cite{saito2021openmatch}               & \multicolumn{1}{c|}{58.17} & \multicolumn{1}{c|}{59.82}  & \multicolumn{1}{c|}{56.71} & 79.46   \\ \hline
      Ours              & \multicolumn{1}{c|}{\textbf{67.99}}      & \multicolumn{1}{c|}{\textbf{66.99}}       & \multicolumn{1}{c|}{\textbf{65.32}}      & \textbf{83.21}        \\ \hline  \hline
    \end{tabular}
  \end{minipage}%
  \hfill
  \begin{minipage}[c]{0.46\textwidth}
    \captionof{table}{Ablation study results.\label{table_ablation}}
    \centering
    \begin{tabular}{ccccc}
      \hline \hline
      \multicolumn{1}{c|}{Datasets}                     & \multicolumn{1}{c|}{ISIC} & \multicolumn{1}{c|}{D7pt} & \multicolumn{1}{c|}{PAD} & DN \\ \hline
      \multicolumn{1}{c|}{w/o SSL} & \multicolumn{4}{c}{69.2}                                                                                   \\ \hline
      \multicolumn{1}{c|}{w/o DOE} & \multicolumn{1}{c|}{68.7}          & \multicolumn{1}{c|}{68.2}  & \multicolumn{1}{c|}{67.9}            &  66.0     \\ \hline
      \multicolumn{1}{c|}{w/o CDS} & \multicolumn{1}{c|}{55.6}          & \multicolumn{1}{c|}{59.3}        & \multicolumn{1}{c|}{61.1}            & 51.9        \\ \hline
      \multicolumn{1}{c|}{w/o DA}  & \multicolumn{1}{c|}{70.5}          & \multicolumn{1}{c|}{69.3}        & \multicolumn{1}{c|}{68.5}            & 65.6        \\ \hline
      \multicolumn{1}{c|}{Ours}    & \multicolumn{1}{c|}{71.1}      & \multicolumn{1}{c|}{70.9}    & \multicolumn{1}{c|}{70.8}        & 69.6    \\ \hline  \hline
    \end{tabular}
  \end{minipage}
\end{table}

\begin{figure}[t]
    \centering
    \includegraphics[width=11cm]{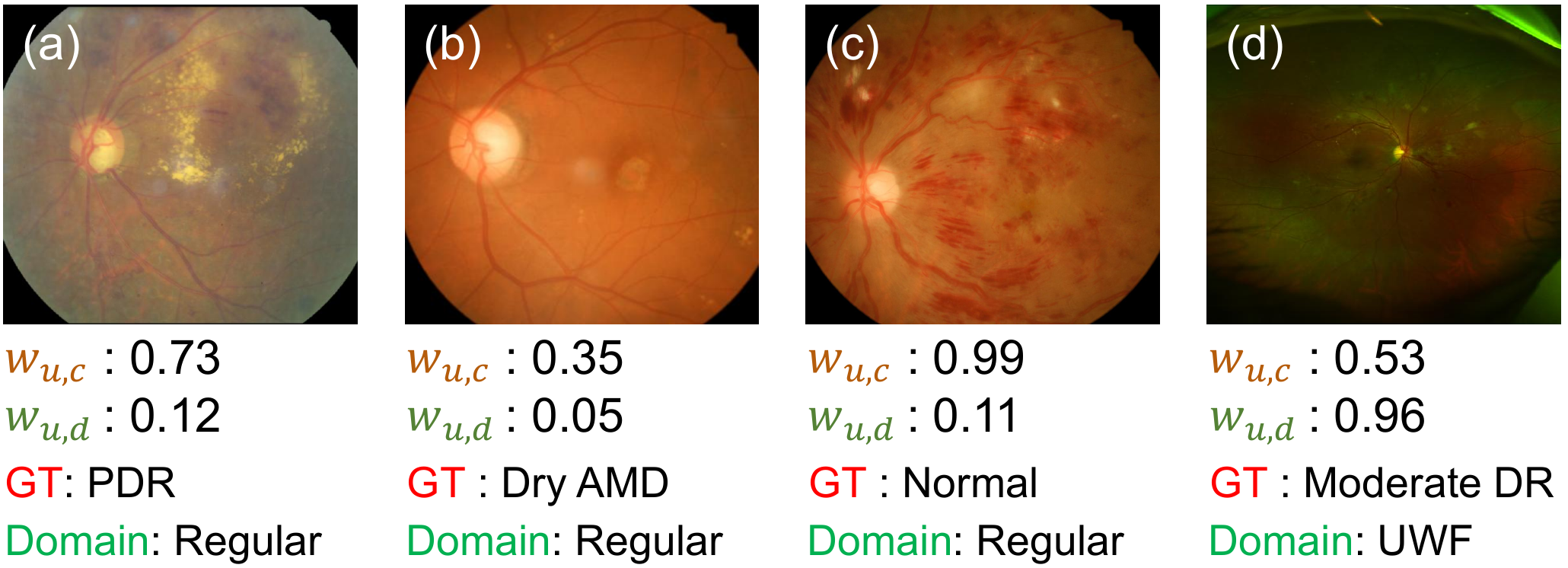}
    \caption{The visualized examples from unlabeled data with normalized scores.}
    \label{fig:vis}
\end{figure}

\subsection{Ablation Study}
\noindent \textbf{Analysis on the Components} We perform an ablation study on each component, and the results are shown in Table \textcolor{red}{3}. The 'w/o DOE' denotes that we use the unlabeled data re-weighed by $w_{d}$ to train the $\Pi$-model without considering the class mismatch. The 'w/o CDS' denotes that we directly take the whole unlabeled data as the unknown domains for domain adaptation without considering the condition of domain mix, \textit{e.g.,} ISIC \& Dermnet. The 'w/o DA' denotes that we exploit $w_{c}$ and $w_{d}$ to re-weight the unlabeled samples but without the adversarial training term. It is found that the dataset with a larger domain gap such as Dermnet suffers from more performance degradation (from 69.6\% to 65.6\%).

\noindent \textbf{Semantic Correlations of Open-set Samples} To explore semantic correlations between samples exhibiting class/domain mismatch, we visualize instances from unlabeled data alongside their corresponding normalized UKC/UKD scores, as depicted in Fig.~\ref{fig:vis}. It is noteworthy that while Fig.~\ref{fig:vis}-(c) exhibits no AMD-related lesions, it yields the highest $w_{u,c}$, indicating a higher semantic similarity in the feature space, e.g., hemorrhages. Incorporating such samples into unified training can effectively enhance model performance by enriching useful features.

\section{Conclusion}
In this work, we propose a novel universal semi-supervised learning framework for medical image classification.
We propose two novel scoring mechanisms for the separation of samples from unknown classes and unknown domains. Then, we adopt regular semi-supervised learning and domain adaptation on the re-weighted unlabeled samples for unified training. Our experimental results show that our proposed methods can perform well under different realistic scenarios.
\\ 
\subsubsection{Acknowledgments.} The work was partially supported by Airdoc medical AI projects donation Phase 2, Monash-Airdoc Research Centre, and in part by the MRFF NCRI GA89126.
\subsubsection{Disclosure of Interests.} The authors declare that they have no competing interests.

\bibliographystyle{splncs04}
\bibliography{Paper-4152}

\begin{thebibliography}{10}
\providecommand{\url}[1]{\texttt{#1}}
\providecommand{\urlprefix}{URL }
\providecommand{\doi}[1]{https://doi.org/#1}

\bibitem{berthelot2019mixmatch}
Berthelot, D., Carlini, N., Goodfellow, I., Papernot, N., Oliver, A., Raffel, C.A.: Mixmatch: A holistic approach to semi-supervised learning. Advances in Neural Information Processing Systems  \textbf{32} (2019)

\bibitem{cao2018partial}
Cao, Z., Ma, L., Long, M., Wang, J.: Partial adversarial domain adaptation. In: European Conference on Computer Vision. pp. 135--150 (2018)

\bibitem{chen2020semi}
Chen, Y., Zhu, X., Li, W., Gong, S.: Semi-supervised learning under class distribution mismatch. In: Proceedings of the AAAI Conference on Artificial Intelligence. vol. 34(4), pp. 3569--3576 (2020)

\bibitem{Dermnet}
Dermnet: Dermnet (2023), \url{https://dermnet.com/}

\bibitem{esteva2017dermatologist}
Esteva, A., Kuprel, B., Novoa, R.A., Ko, J., Swetter, S.M., Blau, H.M., Thrun, S.: Dermatologist-level classification of skin cancer with deep neural networks. Nature  \textbf{542}(7639),  115--118 (2017)

\bibitem{guo2020safe}
Guo, L.Z., Zhang, Z.Y., Jiang, Y., Li, Y.F., Zhou, Z.H.: Safe deep semi-supervised learning for unseen-class unlabeled data. In: International Conference on Machine Learning. pp. 3897--3906. PMLR (2020)

\bibitem{he2016deep}
He, K., Zhang, X., Ren, S., Sun, J.: Deep residual learning for image recognition. In: IEEE/CVF Conference on Computer Vision and Pattern Recognition. pp. 770--778 (2016)

\bibitem{huang2021trash}
Huang, J., Fang, C., Chen, W., Chai, Z., Wei, X., Wei, P., Lin, L., Li, G.: Trash to treasure: harvesting ood data with cross-modal matching for open-set semi-supervised learning. In: IEEE/CVF International Conference on Computer Vision. pp. 8310--8319 (2021)

\bibitem{huang2021universal}
Huang, Z., Xue, C., Han, B., Yang, J., Gong, C.: Universal semi-supervised learning. Advances in Neural Information Processing Systems  \textbf{34},  26714--26725 (2021)

\bibitem{isic}
ISIC: Isic archive (2023), \url{https://www.isic-archive.com/}

\bibitem{ju2021leveraging}
Ju, L., Wang, X., Zhao, X., Bonnington, P., Drummond, T., Ge, Z.: Leveraging regular fundus images for training uwf fundus diagnosis models via adversarial learning and pseudo-labeling. IEEE Transactions on Medical Imaging  \textbf{40}(10),  2911--2925 (2021)

\bibitem{ju2021synergic}
Ju, L., Wang, X., Zhao, X., Lu, H., Mahapatra, D., Bonnington, P., Ge, Z.: Synergic adversarial label learning for grading retinal diseases via knowledge distillation and multi-task learning. IEEE Journal of Biomedical and Health Informatics  \textbf{25}(10),  3709--3720 (2021)

\bibitem{ju2022flexible}
Ju, L., Wu, Y., Wang, L., Yu, Z., Zhao, X., Wang, X., Bonnington, P., Ge, Z.: Flexible sampling for long-tailed skin lesion classification. In: Medical Image Computing and Computer Assisted Intervention--MICCAI 2022. pp. 462--471. Springer (2022)

\bibitem{kawahara2018seven}
Kawahara, J., Daneshvar, S., Argenziano, G., Hamarneh, G.: Seven-point checklist and skin lesion classification using multitask multimodal neural nets. IEEE Journal of Biomedical and Health Informatics  \textbf{23}(2),  538--546 (2018)

\bibitem{laine2016temporal}
Laine, S., Aila, T.: Temporal ensembling for semi-supervised learning. arXiv preprint arXiv:1610.02242  (2016)

\bibitem{lee2013pseudo}
Lee, D.H., et~al.: Pseudo-label: The simple and efficient semi-supervised learning method for deep neural networks. In: ICML Workshop on challenges in representation learning. vol.~3(2), p.~896 (2013)

\bibitem{lee2022contrastive}
Lee, D., Kim, S., Kim, I., Cheon, Y., Cho, M., Han, W.S.: Contrastive regularization for semi-supervised learning. In: IEEE/CVF Conference on Computer Vision and Pattern Recognition. pp. 3911--3920 (2022)

\bibitem{lidividemix}
Li, J., Socher, R., Hoi, S.C.: Dividemix: Learning with noisy labels as semi-supervised learning. In: International Conference on Learning Representations (2020)

\bibitem{liu2020semi}
Liu, Q., Yu, L., Luo, L., Dou, Q., Heng, P.A.: Semi-supervised medical image classification with relation-driven self-ensembling model. IEEE Transactions on Medical Imaging  \textbf{39}(11),  3429--3440 (2020)

\bibitem{ming2022exploit}
Ming, Y., Sun, Y., Dia, O., Li, Y.: How to exploit hyperspherical embeddings for out-of-distribution detection? arXiv preprint arXiv:2203.04450  (2022)

\bibitem{miyato2018virtual}
Miyato, T., Maeda, S.i., Koyama, M., Ishii, S.: Virtual adversarial training: a regularization method for supervised and semi-supervised learning. IEEE Transactions on Pattern Analysis and Machine Intelligence  \textbf{41}(8),  1979--1993 (2018)

\bibitem{pacheco2020pad}
Pacheco, A.G., Lima, G.R., Salomao, A.S., Krohling, B., Biral, I.P., de~Angelo, G.G., Alves~Jr, F.C., Esgario, J.G., Simora, A.C., Castro, P.B., et~al.: Pad-ufes-20: A skin lesion dataset composed of patient data and clinical images collected from smartphones. Data in Brief  \textbf{32},  106221 (2020)

\bibitem{saito2021openmatch}
Saito, K., Kim, D., Saenko, K.: Openmatch: Open-set semi-supervised learning with open-set consistency regularization. Advances in Neural Information Processing Systems  \textbf{34},  25956--25967 (2021)

\bibitem{sindhwani2005co}
Sindhwani, V., Niyogi, P., Belkin, M.: A co-regularization approach to semi-supervised learning with multiple views. In: Proceedings of ICML workshop on learning with multiple views. vol.~2005, pp. 74--79 (2005)

\bibitem{sohn2020fixmatch}
Sohn, K., Berthelot, D., Carlini, N., Zhang, Z., Zhang, H., Raffel, C.A., Cubuk, E.D., Kurakin, A., Li, C.L.: Fixmatch: Simplifying semi-supervised learning with consistency and confidence. Advances in Neural Information Processing Systems  \textbf{33},  596--608 (2020)

\bibitem{sun2020conditional}
Sun, X., Yang, Z., Zhang, C., Ling, K.V., Peng, G.: Conditional gaussian distribution learning for open set recognition. In: Proceedings of the IEEE/CVF Conference on Computer Vision and Pattern Recognition. pp. 13480--13489 (2020)

\bibitem{tarvainen2017mean}
Tarvainen, A., Valpola, H.: Mean teachers are better role models: Weight-averaged consistency targets improve semi-supervised deep learning results. Advances in Neural Information Processing Systems  \textbf{30} (2017)

\bibitem{ye2021towards}
Ye, H., Xie, C., Cai, T., Li, R., Li, Z., Wang, L.: Towards a theoretical framework of out-of-distribution generalization. Advances in Neural Information Processing Systems  \textbf{34},  23519--23531 (2021)

\bibitem{yu2020multi}
Yu, Q., Ikami, D., Irie, G., Aizawa, K.: Multi-task curriculum framework for open-set semi-supervised learning. In: European Conference on Computer Vision. pp. 438--454. Springer (2020)

\end{thebibliography}
\end{document}